\documentclass[letterpaper, 10 pt, conference]{ieeeconf}  % Comment this line out if you need a4paper

\IEEEoverridecommandlockouts                              % This command is only needed if 
\usepackage{graphicx}
\usepackage{caption}
\usepackage{amsmath}
\usepackage{amssymb}
\usepackage[caption=false]{subfig}
\usepackage{booktabs}
\usepackage{multirow}
\usepackage{array}

\usepackage{xcolor}
\definecolor{correction}{RGB}{255, 0, 0}

\definecolor{black}{RGB}{0, 0, 0}
\definecolor{orange}{RGB}{230, 159, 0}
\definecolor{skyblue}{RGB}{86, 180, 233}
\definecolor{green}{RGB}{0, 158, 115}
\definecolor{yellow}{RGB}{240, 228, 66}
\definecolor{blue}{RGB}{0, 114, 178}
\definecolor{red}{RGB}{213, 94, 0}
\definecolor{purple}{RGB}{204, 121, 167}
\definecolor{dark-gray}{RGB}{176,176,176}
\definecolor{light-gray}{RGB}{204,204,204}
\usepackage[utf8]{inputenc}
\usepackage{pgfplots}
\pgfplotsset{compat=1.11,
    /pgfplots/ybar legend/.style={
    /pgfplots/legend image code/.code={%
       \draw[##1,/tikz/.cd,yshift=-0.25em]
        (0cm,0cm) rectangle (3pt,0.8em);},
   },
}
\DeclareUnicodeCharacter{2212}{−}
\usepgfplotslibrary{groupplots,dateplot}
\usepackage{tikz}
\usetikzlibrary{shapes.geometric}
\usetikzlibrary{arrows.meta,arrows}
\usetikzlibrary{patterns,shapes.arrows}
\pgfplotsset{compat=newest}
\usepackage{environ}
\makeatletter
\newsavebox{\measure@tikzpicture}
\NewEnviron{scaletikzpicturetowidth}[1]{%
  \def\tikz@width{#1}%
  \begin{lrbox}{\measure@tikzpicture}%
  \BODY
  \end{lrbox}%
  \pgfmathparse{#1/\wd\measure@tikzpicture}%
  \BODY
}
\makeatother

\usepackage{hyperref}       % hyperlinks
\definecolor{dark-red}{rgb}{0.4,0.15,0.15}
\definecolor{dark-blue}{rgb}{0.15,0.15,0.8}
\definecolor{medium-blue}{rgb}{0,0,0.5}
\hypersetup{
    colorlinks, linkcolor={dark-red},
    citecolor={dark-blue}, urlcolor={medium-blue}
}

\title{\LARGE \bf
Classifying Subjective Time Perception in a Multi-robot Control Scenario Using Eye-tracking Information
}

\author{Till Aust,$^{1}$ Julian Kaduk,$^{1}$ and Heiko Hamann$^{1,2}$% <-this % stops a space
\thanks{$^{1}$Till Aust, Julian Kaduk and Heiko Hamann are with the Department of Computer and Information Science, University of Konstanz, Konstanz, Germany. 
{\tt\small till.aust@uni-konstanz.de}}%
\thanks{$^{2}$Heiko Hamann is member of the 
Centre for the Advanced Study of Collective Behaviour (CASCB), Universität Konstanz, Konstanz, Germany
}
}

\begin{document}

\maketitle
\thispagestyle{empty}
\pagestyle{empty}

%%%%%%%%%%%%%%%%%%%%%%%%%%%%%%%%%%%%%%%%%%%%%%%%%%%%%%%%%%%%%%%%%%%%%%%%%%%%%%%%
\begin{abstract}
%
% Planned Submission: IEEE SMC (Deadline 4.4.25)
%
As automation and mobile robotics reshape work environments, rising expectations for productivity increase cognitive demands on human operators, leading to potential stress and cognitive overload.
%As automation and robotics reshape work environments, cognitive demands on human operators increase, leading to potential stress and cognitive overload. 
%With the rise of automation and mobile robots, work environments change quickly and mental health of the workforce is increasingly challenged. 
Accurately assessing an operator's mental state is critical for maintaining performance and well-being. 
We use subjective time perception, which can be altered by stress and cognitive load, as a sensitive, low-latency indicator of well-being and cognitive strain. 
%Subjective time perception, which is known to be influenced by stress and cognitive load, offers a promising indicator of cognitive strain. 
Distortions in time perception can affect decision-making, reaction times, and overall task effectiveness, making it a valuable metric for adaptive human-swarm interaction systems. 

We study how human physiological signals can be used to estimate a person's subjective time perception in a human-swarm interaction scenario as example. 
A~human operator needs to guide and control a swarm of small mobile robots. 
We obtain eye-tracking data that is classified for subjective time perception based on questionnaire data. 
%We use automated machine learning  for model fine-tuning and SHAP~analysis to explain feature importance. 
Our results show that we successfully estimate a person's time perception from eye-tracking data. 
The approach can profit from individual-based pretraining using only 30~seconds of data. 
%The automatic classification of physiological data will allow us to let the robots react to the human operator's needs in a control loop. 
In future work, we aim for robots that respond to human operator needs by automatically classifying physiological data in a closed control loop. 
%Our research will allow for more scalable and efficient human oversight of large-scale multi-robot systems in the future and generally help to improve human well-being in semi-automatized work environments.
\end{abstract}

%%%%%%%%%%%%%%%%%%%%%%%%%%%%%%%%%%%%%%%%%%%%%%%%%%%%%%%%%%%%%%%%%%%%%%%%%%%%%%%%
\section{Introduction}
% some introduction what is time perception why is it useful
In today's rapidly evolving work environments, driven increasingly by automation, prioritizing the mental health of the workforce is becoming essential~\cite{McGrath2023}.
Workplace stressors can negatively impact well-being, decrease productivity, and potentially lead to burnout. 
A~potential indicator is subjective time perception, which may be experienced as slower under stress~\cite{Ogden2020}. 
When a person is positively engaged, they tend to perceive time as passing more quickly.

% we can use eye-tracking in our project because it is a promising source of information 
In our project \textit{ChronoPilot}~\cite{Botev2021}, we attempt to actively modulate time perception of workers in virtual~(VR) or augmented reality~(AR) automation scenarios in order to increase their well-being and productivity. 
We aim to develop a device that both adjusts time perception and simultaneously measures the user state by obtaining human feedback.
We modulate users' time perception through haptic~\cite{Cavdan2023}, visual~\cite{Schatzschneider2016} auditory~\cite{Picard2022}, or situational stimuli~\cite{Kaduk2023}. 
A~user's time perception state can be estimated by non-invasively measuring their physiological signals~\cite{Aust2024,Orlandic2021}. 
A~promising physiological signal is eye-tracking due to its great availability, minimal intrusiveness, and reliable measurement across a variety of settings~\cite{Chen2016}. 
Many current and future tasks/jobs can be performed or supported using VR or AR devices, such as the Vision Pro\footnote{\tiny\url{https://www.apple.com/de/apple-vision-pro/}} or Meta Quest~3.\footnote{\tiny\url{https://www.meta.com/de/quest/quest-3/}} 
These devices capture eye-tracking data as part of their functionality, hence, requiring no additional hardware. 
% A task can be swarm robotics 
One such task, considering the rise of mobile robots, multi-robot systems, and swarm robotics~\cite{Hamann2018,Dorigo2021}, could be to guide or control semi-autonomous groups of robots.
This human-robot interaction or even more so human-swarm interaction~\cite{Chen2010,Divband2021} may have high demands for a human operator and may potentially be stressful. 
% influence of the robot swarm; well-being & subjective time perception
Research has shown that passively observing~\cite{Podevijn2016} or controlling~\cite{Kaduk2024} a larger number of robots moving around, and also smoothness in robot motion~\cite{Dietz2017} can influence the participant's well-being.
%Research has shown that already passively observing a larger number of robots moving around a participant can results in higher self-reported levels of arousal~\cite{Podevijn2016}. 
%Not only the number of robots can affect the mental well-being but also the smoothness in robot motion, increasing emotional valence or faster movement, increasing arousal, affects the human well-being when observing a swarm of robots~\cite{Dietz2017}. 
% time perception in robot swarm 
%Similar to the experiences of air traffic controllers~\cite{Aust2024}, also the operator of a robot swarm may experience changes in their subjective time perception. 
%This may influence their well-being and decision-making capabilities. 
Participants reported a faster passage of time when controlling a larger number of robots~\cite{Kaduk2023}. 
% Previous research has shown a faster passage of time perception and increased perception of flow when being asked to supervise 15 robots in a swarm compared to 1 or 5~\cite{Kaduk2023}. 
% The work in this paper is based on the eye tracking data collected from a similar study in which the the total number of robots was kept constant with~$N=15$ robots, however only an active subset of $N_{\text{a}} \in \{1, 3, 5, 7, 9, 11, 13, 15\}$ robots was moving and therefore had to be controlled~\cite{Kaduk2024}. 
% This study revealed that number of active robot only significantly affected time perception between one ore more than on robot being active, while the level of arousal and the perceived task difficulty increased with an increasing number of active robots. 

% Feeback for robot swarms 
Possible approaches to moderating stressful effects of controlling a robot swarm include the use of automated after-action reviews~\cite{Qian2024} or monitoring the participants' heart rate variability (HRV)~\cite{Villani2020}. % to estimate their mental fatigue and adapt the robot behavior accordingly~\cite{Villani2020}.
%can be used, to control the robots and to keep users in good ranges of well-being and mental workload~. 
In automated after-action reviews, the human-robot task is replayed to the participant directly after execution and questionnaires need to be filled. 
This is not applicable in our case, as we aim to monitor the time perception online in order to actively modify the work environment. 
Moreover, interrupting the task to review time perception may skew the results as by disturbing the participant's workflow. 
Instead, measurements should occur in parallel with the participant's task, ideally without the participant's conscious awareness. 
% eye-tracking as information source and as feeback
We prefer to use the participant's physiological data, such as HRV and eye-tracking. 
A~common approach to process eye-tracking data is to extract features, such as the pupil size, eye movement, or blinking. 
These are compiled into a feature vector and classified using machine learning (ML) algorithms~\cite{Lim2022}. 
% examples of application 
Utilizing this approach, eye-tracking information can be used to detect stress~\cite{Tao2022}, monitor mental workload~\cite{Guo2021}, or recognize emotion~\cite{Lim2020}. 
%Tao~\textit{et al.}~\cite{Tao2022} used electrodermal activity, heart rate, and eye-blink rate to do stress detection using machine learning in a VR~environment. 
%They employ a background subtraction on the eye-tracking data and obtain an accuracy of 74.2~\% for the 3 classes no stress, low stress, and high stress.
%Guo~\textit{et al.}~\cite{Guo2021} used the eye-tracking information of operators of space telerobotics during training sequences to monitor the mental workload. 
%They found that fixation duration, saccade frequency and duration, pupil diameter, and index of pupillary activity are significant features. 
% Lim~\textit{et al.}~\cite{Lim2020} reviews approaches doing emotion recognition based on eye-tracking data. 
% They report that so far there are only a limited number of studies in this field where typically eye-tracking was paired with other data sources (e.g., Electroencephalography).
% They curated the following list of emotional-relevant features: pupil diameter, Electrooculography (EOG), pupil position, fixation duration of the eye, distance between sclera and iris, motion speed of the eye, and pupillary responses.

% automated machine learning 
A common limitation of these studies is the ad hoc selection of ML models, with no clear rationale for choosing one over another.
Model selection often seems to rely on manual parameter tuning or comparisons within a limited set of models, lacking a systematic approach.
Using automated machine learning (AutoML) is a potential solution providing justification and ensuring thorough exploration of search spaces for ML models and their hyperparameter configurations. 
AutoML methods automatically compose ML algorithms into a pipeline and choose appropriate hyperparameters to optimize a given metric~\cite{Yao2018}.  
Existing frameworks like \textit{auto-sklearn}~\cite{Feurer2015,Feurer2020} and GAMA~\cite{Gijsbers2019} provide ready-to-use implementations. 
More recently, \textit{Mohr}~and~\textit{Wever}~\cite{Mohr2022} introduced the Naive AutoML framework, which achieves comparable performance with a substantial reduction in computation time. 
In this paper, we study how to derive subjective time perception feedback from eye-tracking data obtained during a multi-robot controlling scenario, using an automated ML approach based on Naive AutoML~\cite{Mohr2022}, with the goal of closing the feedback loop of the \textit{ChronoPilot} device. 
% We propose an automated machine learning approach based on Naive AutoML~\cite{Mohr2022} to find an effective and fine-tuned machine learning pipeline for classifying subjective time perception based on eye-tracking features.

\section{Experiments and Methodology}
The eye-tracking data were collected in robot experiments with participants controlling a robot swarm~\cite{Kaduk2024}. 
We preprocess the eye-tracking data for an AutoML approach to optimize an ML pipeline and its hyperparameters. 

\subsection{Robot experiment}
\begin{figure}[t]
    \centering
    \vspace{5pt}
    \includegraphics[height=0.45\linewidth]{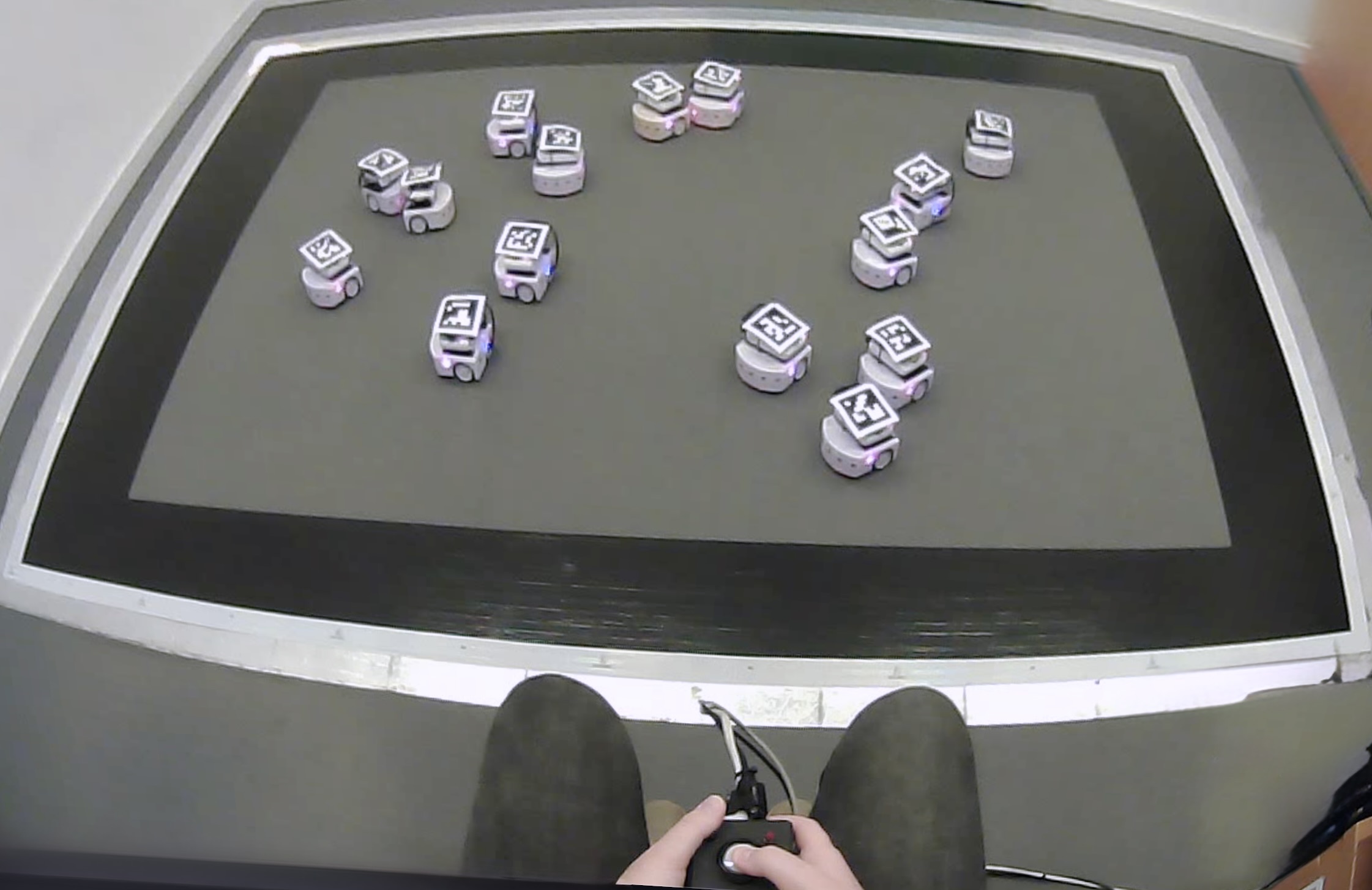}
    \caption{POV of participant during the robot experiment.}
    \label{fig:pov_experiment}
    \vspace{-10pt} % Removes extra space
\end{figure}
Participants control a swarm of small mobile robots (Thymio~II~\cite{Riedo2013}) using a single-button interface (see Fig.~\ref{fig:pov_experiment}) across 24 randomized trials, while eye-tracking and questionnaires capture physiological responses and perceived time during varying trial durations and counts of active robots. 
A more detailed description is given by Kaduk~\textit{et al.}~\cite{Kaduk2024}. 
%Hence, we only give a short summary consisting of the most relevant information. 
%The full study is presented in detail in a previous paper evaluating the self reported measures and psychological effects on the participants~. 
%The following therefore only presents a brief summary. 

In the experiment, we differentiate between active and passive robots. 
The active robots follow a random walk with obstacle avoidance and user input, while the passive robots are stationary at random locations. 
The participant has a single-button interface with the task to keep all active robots within a rectangle (2.2~m by 1.6~m) marked by black tape on the floor. 
When the button is pressed, all active robots rotate on the spot in random direction until either the button is released or a 3-second timer has elapsed.
The experiment was split into 24 different trials randomized in order for each participant. 
The two variable parameters were the trial duration of either 1~min, 3~min, or 5~min and the number of actively moving robots $N_\text{active} \in \{1, 3, \dots, 15\}$. Before each trial, the participant was asked to watch a relaxing aquatic video for two minutes to provide a neutral baseline of the physiological data~\cite{Piferi2000}. Each participant completed these trials during two appointments of approximately 1.5~hours each.
The participants were equipped with a Pupil Core eye-tracking platform.\footnote{\tiny\url{https://pupil-labs.com/products/core}} 
After each trial, the participant filled in a questionnaire including estimates of their subjective perceived passage of time (PPOT) on a 5-point Likert scale and the duration of the experiment on a visual timeline ranging from 0:00~to~10:00~minutes. 

\subsection{Eye-tracking data}
All experiment data were recorded with synchronized timestamps using the lab steaming layer~\cite{Kothe2024} and the Pupil Labs software (also used for calibration of the Pupil Core eye-tracking device). 
The timestamps were used to split the data into individual trial sequences.
%The initial preprocessing, for example, aligning the eye-tracking data to the experiment and extracting the relevant sequences was done using the pupil labs software. 
%\ta{TODO @JK: any preprocessing using the pupils software? Did I forget something?}
A total of 21 participants completed the 24 trials each, yielding 504~experiment sequences. 
%Additionally, the participants were asked to watch a relaxing aquatic video before each trial~\cite{Piferi2000}, giving us additional baseline sequences for each trial sequence. 
Due to software malfunctions and calibration losses, 99~sequences (19.6\%) were excluded from analysis. 

%We have a theoretical total of 504~experiment sequences (21 participants $\times$ 24 experimental settings) of varying length depending on the experimental length parameter. 
%Additionally, before each experiment we have collected a baseline sequence of approximately 2~minutes. 
All sequences contain the normalized x and y position of the pupil center, the 2-dimensional pupil diameter for the left and right eye, the 3-dimensional pupil diameter for the left and right eye, the fixation ID (indicating the number of fixations), the normalized x and y position of the fixation, the dispersion, and the duration recorded by the Pupil Core eye-tracking platform. 
%However, we had to exclude 99~experiment sequences (19.6~\% of the data) due to data corruption \ta{or participants not completing all experiment settings @JK koennen wir das so sagen?}.
% DISTRIBUTION OF MISSING DATA (total: 195 sequences)
%  Times: 1 min - 64; 3 min - 65; 5 min - 66
%  Robots: 1  3  5  7  9 11 13 15
%  miss   24 23 24 25 26 24 22 27
%  Part: 1  2  3  4  5  6  7  8  9  10 11 12 13 14 15 16 17 18 19 20 21 22 23 24 25
%  miss 12 12 13  1  5  1  1 24 24   2  2  2  1  2  1  2 14  7  1 15  2  1  2 24 24
% -> somewhat balanced missing 

The remaining 405~experiment eye-tracking sequences are divided into non-overlapping slices of $t_w$ = \{1~s, 2~s, 5~s, 10~s, 15~s, 20~s, 30~s, 45~s, 60~s\} (extending the time window selection of~\cite{Guo2021}). 
Subsequently, the sliced sequences and the baseline sequences are used to calculate the features. 

From the sequences, we extract 26~features that are grouped into features related to eye movement (total of 10~features) and features related to pupillary response (total of 16~features). 
The features related to eye movement are: fixation frequency, fixation duration mean, fixation duration maximum, fixation dispersion mean, fixation dispersion maximum, saccade frequency, saccade duration mean, saccade duration maximum, saccade speed mean, and saccade speed maximum.  
The features related to pupillary response are: pupil diameter 2d mean/maximum/standard deviation for the left and right eye, pupil diameter 3d mean/maximum/standard deviation for the left and right eye, and index of pupillary activity~(IPA)~\cite{Duchowski2018} derived from 2d and 3d for the left and right eye. 
% right eye = 0; left eye = 1

We apply baseline subtraction to all features calculated from the experiment sequences. 
We subtract the feature value derived from the baseline condition from the corresponding feature in the experiment data slice. 
This adjustment compensates for potential variation in illumination due to daytime or lamp settings, which can significantly impact the eye-tracking features~\cite{Chen2016} differently for each participant. 

Finally, we split the dataset into an analysis dataset, which is used for training and cross-validation, and a test dataset which is used for the final performance evaluation. 
To ensure reproducibility, we have made the code publicly available.\footnote{\tiny\url{https://github.com/tilly111/chronopilot_feature_extraction}}\footnote{\tiny\url{https://github.com/tilly111/chronopilot_classification}}

\subsection{Time perception labels}
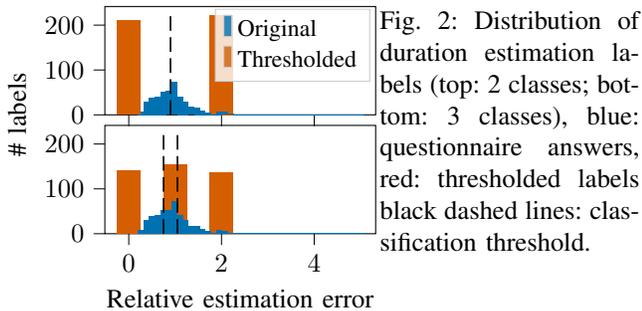
\begin{figure}[t]%
    \centering
    \vspace{5pt}
    \begin{minipage}{0.5\linewidth}
        % This file was created with tikzplotlib v0.10.1.
\begin{tikzpicture}%[show background rectangle]

\pgfplotsset{
    legend image code/.code={
        \draw [#1] (0cm,-0.1cm) rectangle (0.6cm,0.1cm);
    },
}
%\draw[red, dashed] (current bounding box.south west) rectangle (current bounding box.north east);
\begin{groupplot}[%group style={group size=1 by 2}
group style={
        group size=1 by 2,
        xlabels at=edge bottom,
        ylabels at=edge left,
        xticklabels at=edge bottom,
        vertical sep=4pt
    },
    width=8cm,
    height=3.5cm,
    xlabel={Relative estimation error},
    xmin=0, xmax=7,
    tickpos=left
]
\nextgroupplot[
legend cell align={left},
legend style={fill opacity=0.8, draw opacity=1, text opacity=1, draw=light-gray, legend style={font=\small}},
tick align=outside,
tick pos=left,
x grid style={dark-gray},
xmin=-0.5155, xmax=5.3255,
xtick style={color=black},
y grid style={dark-gray},
ymin=0, ymax=241.5,
ytick style={color=black},
width=1.2\linewidth,
height=0.7\linewidth
]
\draw[draw=none,fill=red] (axis cs:1.75,0) rectangle (axis cs:2.25,222);
\draw[draw=none,fill=blue] (axis cs:0.198,0) rectangle (axis cs:0.31955,7);
\addlegendimage{ybar,ybar legend,draw=none,fill=blue}
\addlegendentry{Original}

\draw[draw=none,fill=red] (axis cs:-0.25,0) rectangle (axis cs:0.25,210);
\addlegendimage{ybar,ybar legend,fill=red}
\addlegendentry{Thresholded}

\draw[draw=none,fill=blue] (axis cs:0.31955,0) rectangle (axis cs:0.4411,30);
\draw[draw=none,fill=blue] (axis cs:0.4411,0) rectangle (axis cs:0.56265,38);
\draw[draw=none,fill=blue] (axis cs:0.56265,0) rectangle (axis cs:0.6842,41);
\draw[draw=none,fill=blue] (axis cs:0.6842,0) rectangle (axis cs:0.80575,52);
\draw[draw=none,fill=blue] (axis cs:0.80575,0) rectangle (axis cs:0.9273,53);
\draw[draw=none,fill=blue] (axis cs:0.9273,0) rectangle (axis cs:1.04885,73);
\draw[draw=none,fill=blue] (axis cs:1.04885,0) rectangle (axis cs:1.1704,40);
\draw[draw=none,fill=blue] (axis cs:1.1704,0) rectangle (axis cs:1.29195,30);
\draw[draw=none,fill=blue] (axis cs:1.29195,0) rectangle (axis cs:1.4135,17);
\draw[draw=none,fill=blue] (axis cs:1.4135,0) rectangle (axis cs:1.53505,16);
\draw[draw=none,fill=blue] (axis cs:1.53505,0) rectangle (axis cs:1.6566,10);
\draw[draw=none,fill=blue] (axis cs:1.6566,0) rectangle (axis cs:1.77815,5);
\draw[draw=none,fill=blue] (axis cs:1.77815,0) rectangle (axis cs:1.8997,1);
\draw[draw=none,fill=blue] (axis cs:1.8997,0) rectangle (axis cs:2.02125,7);
\draw[draw=none,fill=blue] (axis cs:2.02125,0) rectangle (axis cs:2.1428,6);
\draw[draw=none,fill=blue] (axis cs:2.1428,0) rectangle (axis cs:2.26435,2);
\draw[draw=none,fill=blue] (axis cs:2.26435,0) rectangle (axis cs:2.3859,0);
\draw[draw=none,fill=blue] (axis cs:2.3859,0) rectangle (axis cs:2.50745,1);
\draw[draw=none,fill=blue] (axis cs:2.50745,0) rectangle (axis cs:2.629,0);
\draw[draw=none,fill=blue] (axis cs:2.629,0) rectangle (axis cs:2.75055,0);
\draw[draw=none,fill=blue] (axis cs:2.75055,0) rectangle (axis cs:2.8721,0);
\draw[draw=none,fill=blue] (axis cs:2.8721,0) rectangle (axis cs:2.99365,1);
\draw[draw=none,fill=blue] (axis cs:2.99365,0) rectangle (axis cs:3.1152,0);
\draw[draw=none,fill=blue] (axis cs:3.1152,0) rectangle (axis cs:3.23675,0);
\draw[draw=none,fill=blue] (axis cs:3.23675,0) rectangle (axis cs:3.3583,0);
\draw[draw=none,fill=blue] (axis cs:3.3583,0) rectangle (axis cs:3.47985,0);
\draw[draw=none,fill=blue] (axis cs:3.47985,0) rectangle (axis cs:3.6014,0);
\draw[draw=none,fill=blue] (axis cs:3.6014,0) rectangle (axis cs:3.72295,0);
\draw[draw=none,fill=blue] (axis cs:3.72295,0) rectangle (axis cs:3.8445,0);
\draw[draw=none,fill=blue] (axis cs:3.8445,0) rectangle (axis cs:3.96605,0);
\draw[draw=none,fill=blue] (axis cs:3.96605,0) rectangle (axis cs:4.0876,1);
\draw[draw=none,fill=blue] (axis cs:4.0876,0) rectangle (axis cs:4.20915,0);
\draw[draw=none,fill=blue] (axis cs:4.20915,0) rectangle (axis cs:4.3307,0);
\draw[draw=none,fill=blue] (axis cs:4.3307,0) rectangle (axis cs:4.45225,0);
\draw[draw=none,fill=blue] (axis cs:4.45225,0) rectangle (axis cs:4.5738,0);
\draw[draw=none,fill=blue] (axis cs:4.5738,0) rectangle (axis cs:4.69535,0);
\draw[draw=none,fill=blue] (axis cs:4.69535,0) rectangle (axis cs:4.8169,0);
\draw[draw=none,fill=blue] (axis cs:4.8169,0) rectangle (axis cs:4.93845,0);
\draw[draw=none,fill=blue] (axis cs:4.93845,0) rectangle (axis cs:5.06,1);
\path [draw=black, semithick, dash pattern=on 5.55pt off 2.4pt]
(axis cs:0.9,0)
--(axis cs:0.9,230);

\nextgroupplot[
% legend cell align={left},
% legend style={fill opacity=0.8, draw opacity=1, text opacity=1, draw=lightgray204},
tick align=outside,
tick pos=left,
x grid style={dark-gray},
xmin=-0.5155, xmax=5.3255,
xtick style={color=black},
y grid style={dark-gray},
ymin=0, ymax=241.5,
ytick style={color=black},
ylabel=\# labels,
every axis y label/.append style={at=(ticklabel cs:1.1)},
width=1.2\linewidth,
height=0.7\linewidth
]
\draw[draw=none,fill=red] (axis cs:-0.25,0) rectangle (axis cs:0.25,141);
\draw[draw=none,fill=red] (axis cs:0.75,0) rectangle (axis cs:1.25,155);
\draw[draw=none,fill=red] (axis cs:1.75,0) rectangle (axis cs:2.25,136);
\draw[draw=none,fill=blue] (axis cs:0.198,0) rectangle (axis cs:0.31955,7);
\draw[draw=none,fill=blue] (axis cs:0.31955,0) rectangle (axis cs:0.4411,30);
\draw[draw=none,fill=blue] (axis cs:0.4411,0) rectangle (axis cs:0.56265,38);
\draw[draw=none,fill=blue] (axis cs:0.56265,0) rectangle (axis cs:0.6842,41);
\draw[draw=none,fill=blue] (axis cs:0.6842,0) rectangle (axis cs:0.80575,52);
\draw[draw=none,fill=blue] (axis cs:0.80575,0) rectangle (axis cs:0.9273,53);
\draw[draw=none,fill=blue] (axis cs:0.9273,0) rectangle (axis cs:1.04885,73);
\draw[draw=none,fill=blue] (axis cs:1.04885,0) rectangle (axis cs:1.1704,40);
\draw[draw=none,fill=blue] (axis cs:1.1704,0) rectangle (axis cs:1.29195,30);
\draw[draw=none,fill=blue] (axis cs:1.29195,0) rectangle (axis cs:1.4135,17);
\draw[draw=none,fill=blue] (axis cs:1.4135,0) rectangle (axis cs:1.53505,16);
\draw[draw=none,fill=blue] (axis cs:1.53505,0) rectangle (axis cs:1.6566,10);
\draw[draw=none,fill=blue] (axis cs:1.6566,0) rectangle (axis cs:1.77815,5);
\draw[draw=none,fill=blue] (axis cs:1.77815,0) rectangle (axis cs:1.8997,1);
\draw[draw=none,fill=blue] (axis cs:1.8997,0) rectangle (axis cs:2.02125,7);
\draw[draw=none,fill=blue] (axis cs:2.02125,0) rectangle (axis cs:2.1428,6);
\draw[draw=none,fill=blue] (axis cs:2.1428,0) rectangle (axis cs:2.26435,2);
\draw[draw=none,fill=blue] (axis cs:2.26435,0) rectangle (axis cs:2.3859,0);
\draw[draw=none,fill=blue] (axis cs:2.3859,0) rectangle (axis cs:2.50745,1);
\draw[draw=none,fill=blue] (axis cs:2.50745,0) rectangle (axis cs:2.629,0);
\draw[draw=none,fill=blue] (axis cs:2.629,0) rectangle (axis cs:2.75055,0);
\draw[draw=none,fill=blue] (axis cs:2.75055,0) rectangle (axis cs:2.8721,0);
\draw[draw=none,fill=blue] (axis cs:2.8721,0) rectangle (axis cs:2.99365,1);
\draw[draw=none,fill=blue] (axis cs:2.99365,0) rectangle (axis cs:3.1152,0);
\draw[draw=none,fill=blue] (axis cs:3.1152,0) rectangle (axis cs:3.23675,0);
\draw[draw=none,fill=blue] (axis cs:3.23675,0) rectangle (axis cs:3.3583,0);
\draw[draw=none,fill=blue] (axis cs:3.3583,0) rectangle (axis cs:3.47985,0);
\draw[draw=none,fill=blue] (axis cs:3.47985,0) rectangle (axis cs:3.6014,0);
\draw[draw=none,fill=blue] (axis cs:3.6014,0) rectangle (axis cs:3.72295,0);
\draw[draw=none,fill=blue] (axis cs:3.72295,0) rectangle (axis cs:3.8445,0);
\draw[draw=none,fill=blue] (axis cs:3.8445,0) rectangle (axis cs:3.96605,0);
\draw[draw=none,fill=blue] (axis cs:3.96605,0) rectangle (axis cs:4.0876,1);
\draw[draw=none,fill=blue] (axis cs:4.0876,0) rectangle (axis cs:4.20915,0);
\draw[draw=none,fill=blue] (axis cs:4.20915,0) rectangle (axis cs:4.3307,0);
\draw[draw=none,fill=blue] (axis cs:4.3307,0) rectangle (axis cs:4.45225,0);
\draw[draw=none,fill=blue] (axis cs:4.45225,0) rectangle (axis cs:4.5738,0);
\draw[draw=none,fill=blue] (axis cs:4.5738,0) rectangle (axis cs:4.69535,0);
\draw[draw=none,fill=blue] (axis cs:4.69535,0) rectangle (axis cs:4.8169,0);
\draw[draw=none,fill=blue] (axis cs:4.8169,0) rectangle (axis cs:4.93845,0);
\draw[draw=none,fill=blue] (axis cs:4.93845,0) rectangle (axis cs:5.06,1);
\path [draw=black, semithick, dash pattern=on 5.55pt off 2.4pt]
(axis cs:0.75,0)
--(axis cs:0.75,230);

\path [draw=black, semithick, dash pattern=on 5.55pt off 2.4pt]
(axis cs:1.05,0)
--(axis cs:1.05,230);

\end{groupplot}
% for bounding box of the figure; TODO remove
% \draw[red, dashed] (current bounding box.south west) rectangle (current bounding box.north east);
\end{tikzpicture}%
%
    \end{minipage}%
    \hfill%
    \begin{minipage}{0.4\linewidth}
        \captionsetup{width=1\linewidth}
        \setlength{\abovecaptionskip}{45pt} % Space above caption
        \captionof{figure}{Distribution of duration estimation labels (top: 2~classes; bottom: 3~classes), blue: questionnaire answers, red:  thresholded labels black dashed lines: classification threshold.}
        \label{fig:duration_estimate_labels}
    \end{minipage}
    %\vspace{-45pt} % Removes extra space
    \vspace{-33pt}
\end{figure}%
The participants' answers to the questionnaires after each experiment provide information to label the dataset. 
We have two labels ``duration estimation," derived from the subjective estimation of time, and subjective perceived passage of time (PPOT) which is estimated based on the questionnaire.

We either split the labels into a binary classification problem~\cite{Aust2024} or a three-class problem~\cite{Orlandic2021}. 
The binary classification is an essential benchmark that provides minimal feedback in potential applications. 
The three-class classification problem is interesting for our application of the \textit{ChronoPilot} device. 
It provides all information  (e.g., speed up, slow down, or keep as is) to decide whether and which stimulus needs to be applied to change the time perception of the user. 

\subsubsection{Duration estimate}
After each experiment, participants estimated the duration of the experimental interaction on a visual timeline ranging from 0:00~to~10:00~minutes. 
The estimated duration was then divided by the actual experiment length (e.g., 1, 3, or 5~min) to compute the relative estimation error~$e_\text{rel}$. 
For binary classification, we thresholded the relative estimation error~$e_\text{rel}$ at 0.9 categorizing values $e_\text{rel}\leq 0.9$ as underestimation and $e_\text{rel}> 0.9$ as overestimation. 
This threshold accounts for the general tendency toward underestimation as described by Vierodt's law~\cite{Lejeune2009}. 
For three-class classification, we used thresholds of 0.75~and~1.05: $e_\text{rel}<0.75$ is underestimation, $0.75\leq e_\text{rel} \leq 1.05$ is interpreted as correct estimation, and $e_\text{rel}>1.05$ is overestimation. 
Again, these thresholds address the general trend of underestimation. 
The distribution of labels is shown in Fig.~\ref{fig:duration_estimate_labels}. 
% \begin{figure}
%     \centering
%     % \includegraphics[width=1\linewidth]{figures/duration_estimate_labels.png}
%     \include{figures/duration_estimate_labels}
%     \setlength{\abovecaptionskip}{-30pt} % Space above caption
%     \setlength{\belowcaptionskip}{-10pt} % Space below caption
%     \caption{Distribution of the duration estimation labels (top: 2~classes; bottom: 3~classes). Blue indicates the original questionnaire answers, while red are the thresholded labels. The black dotted lines indicate the classification threshold.}
%     \label{fig:duration_estimate_labels}
% \end{figure}

\subsubsection{Subjective perceived passage of time}
\begin{figure}[t]
    \centering
    \vspace{5pt}
    \begin{minipage}{0.5\linewidth}
        % This file was created with tikzplotlib v0.10.1.
\begin{tikzpicture}

\begin{groupplot}[
group style={
        group size=1 by 2,
        xlabels at=edge bottom,
        ylabels at=edge left,
        xticklabels at=edge bottom,
        vertical sep=4pt
    },
    width=8cm,
    height=3.5cm,
    xlabel={PPOT},
    xlabel style={yshift=10pt},
    xtick={0,1,2,3,4},
    xticklabel style={rotate=45, xshift=-5pt, yshift=15pt},
    xticklabels={very slow,slow,medium,fast,very fast},
    xmin=0, xmax=7,
    tickpos=left
]
\nextgroupplot[
legend cell align={left},
legend style={fill opacity=0.8, draw opacity=1, text opacity=1, draw=light-gray, legend style={font=\small}},
tick align=outside,
tick pos=left,
x grid style={dark-gray},
xmin=-0.365, xmax=4.365,
% xtick style={color=black},
% xtick={0,1,2,3,4},
% xticklabel style={rotate=30.0},
% xticklabels={very slow,slow,medium,fast,very fast},
y grid style={dark-gray},
ymin=0, ymax=241.5,
ytick style={color=black},
width=1.2\linewidth,
height=0.7\linewidth
]
\draw[draw=none,fill=blue] (axis cs:-0.15,0) rectangle (axis cs:0.15,22);
\addlegendimage{ybar,ybar legend,draw=none,fill=blue}
\addlegendentry{Original}

\draw[draw=none,fill=blue] (axis cs:0.85,0) rectangle (axis cs:1.15,107);
\draw[draw=none,fill=blue] (axis cs:1.85,0) rectangle (axis cs:2.15,96);
\draw[draw=none,fill=blue] (axis cs:2.85,0) rectangle (axis cs:3.15,150);
\draw[draw=none,fill=blue] (axis cs:3.85,0) rectangle (axis cs:4.15,57);
\draw[draw=none,fill=red] (axis cs:1.25,0) rectangle (axis cs:1.75,225);
\addlegendimage{ybar,ybar legend,draw=none,fill=red}
\addlegendentry{Thresholded}

\draw[draw=none,fill=red] (axis cs:2.25,0) rectangle (axis cs:2.75,207);
\path [draw=black, semithick, dash pattern=on 5.55pt off 2.4pt]
(axis cs:2.2,0)
--(axis cs:2.2,230);

\nextgroupplot[
legend cell align={left},
legend style={
  fill opacity=0.8,
  draw opacity=1,
  text opacity=1,
  at={(0.03,0.97)},
  anchor=north west,
  draw=light-gray, 
  legend style={font=\scriptsize}
},
tick align=outside,
tick pos=left,
x grid style={dark-gray},
xlabel={PPOT},
xmin=-0.365, xmax=4.365,
xtick style={color=black},
y grid style={dark-gray},
ylabel=\# labels,
every axis y label/.append style={at=(ticklabel cs:1.1)},
ymin=0, ymax=241.5,
ytick style={color=black},
width=1.2\linewidth,
height=0.7\linewidth
]
\draw[draw=none,fill=blue] (axis cs:-0.15,0) rectangle (axis cs:0.15,22);
\draw[draw=none,fill=blue] (axis cs:0.85,0) rectangle (axis cs:1.15,107);
\draw[draw=none,fill=blue] (axis cs:1.85,0) rectangle (axis cs:2.15,96);
\draw[draw=none,fill=blue] (axis cs:2.85,0) rectangle (axis cs:3.15,150);
\draw[draw=none,fill=blue] (axis cs:3.85,0) rectangle (axis cs:4.15,57);
\draw[draw=none,fill=red] (axis cs:0.25,0) rectangle (axis cs:0.75,129);
\draw[draw=none,fill=red] (axis cs:2.05,0) rectangle (axis cs:2.55,96);
\draw[draw=none,fill=red] (axis cs:3.25,0) rectangle (axis cs:3.75,207);
\path [draw=black, semithick, dash pattern=on 5.55pt off 2.4pt]
(axis cs:1.5,0)
--(axis cs:1.5,230);

\path [draw=black, semithick, dash pattern=on 5.55pt off 2.4pt]
(axis cs:2.7,0)
--(axis cs:2.7,230);

\end{groupplot}
% for bounding box of the figure; TODO remove
% \draw[red, dashed] (current bounding box.south west) rectangle (current bounding box.north east);
\end{tikzpicture}%
    \end{minipage}%
    \hfill%
    \begin{minipage}{0.4\linewidth}
        \captionsetup{width=1\linewidth}
        \setlength{\abovecaptionskip}{57pt} % Space above caption
        \captionof{figure}{Distribution of the subjective PPOT labels (top: 2~classes; bottom: 3~classes), blue: questionnaire answers, red: thresholded labels, black dashed lines: classification threshold.}
        \label{fig:ppot_labels}
    \end{minipage}
    \vspace{-45pt} % Removes extra space
\end{figure}
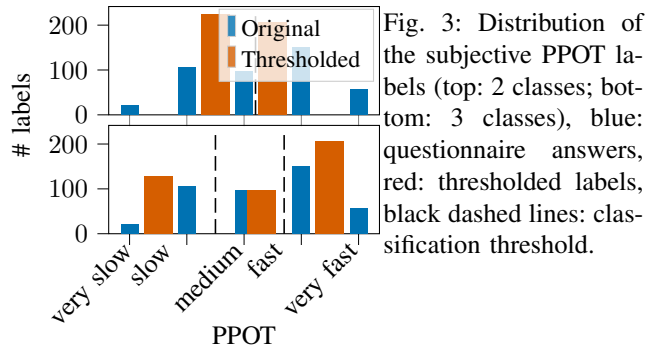

The participants had to estimate their subjective passage of time after each experiment on a 5-point Likert scale ranging from ``very slow" to ``very fast." 
For the binary classification task we thresholded ``very slow" and ``slow" to class slow subjective PPOT and ``neutral," ``fast," and ``very fast" to class fast subjective PPOT (similar to~\cite{Aust2024}). 
For the three class classification problem we combined ``very slow" and ``slow" to class slow subjective PPOT, ``neutral" to class neutral subjective PPOT, and ``fast" and ``very fast" to class fast subjective PPOT. 
An overview of the label distribution is shown in Fig.~\ref{fig:ppot_labels}. 
%
% \begin{figure}
%     \centering
%     % \includegraphics[width=1\linewidth]{figures/ppot_labels.png}
%     \include{figures/ppot_labels}
%     \setlength{\abovecaptionskip}{-30pt} % Space above caption
%     \setlength{\belowcaptionskip}{-10pt} % Space below caption
%     \caption{Distribution of the subjective PPOT labels (top: 2~classes; bottom: 3~classes). Blue indicates the original questionnaire answers, while red are the thresholded labels. The black dotted lines indicate the classification threshold. }
%     \label{fig:ppot_labels}
% \end{figure}

\subsection{Automated machine learning}
We utilize the Naive AutoML library~\cite{Mohr2022} to automatically optimize ML pipelines for our eye-tracking data in a binary or three-class classification task using all eye-tracking features. 
Naive AutoML searches through \textit{sci-kit} learn~\cite{scikit-learn} classifier and preprocessing implementations to identify the classification pipeline that optimizes a given metric. 
This optimization follows a greedy approach in two phases: (1)~identifying the best combination of pre-processors and classifiers with default hyperparameters via enumeration, discarding invalid combinations, and (2)~optimizing the hyperparameters of the selected algorithms using random search (e.g., randomly selecting hyperparameters until a valid configuration is found in the default hyperparameter search space of Naive AutoML). 
After completing a predefined number of optimization steps, the best pipeline found is trained on the full dataset and returned.
Although the greedy approach could potentially result in sub-optimal pipelines, empirical evidence shows that the performance of these solutions typically has little to no gap compared to those obtained by more time-consuming methods~\cite{Mohr2022}.
This library is highly configurable, enabling the setting of timeout limits, maximum hyperparameter iterations, and evaluation metrics.
We do not set a timeout limit but restrict the optimization to 1024 hyperparameter steps and apply early stopping if there is no improvement after 100~iterations.
Further, we excluded the histogram-based gradient boosting classifier as it did not work properly during the hyperparameter iterations. % the hyperparameter iteration was super slow, with factor 1000 or something. 
For evaluation we choose accuracy (ACC), because it will ultimately be the essential measure for our envisioned application and is applicable to both binary classification and three-class classification. 
In all cases, we took the default configuration of 5~independent and stratified 80\%/20\%~training/validation splits of Naive AutoML to evaluate the ML pipelines.

\section{Results and Discussion}
To use eye-tracking data for subjective time perception classification (e.g., feedback control of the \textit{ChronoPilot} device), we do window slicing on the raw eye-tracking data, extract relevant features and use baseline subtraction. 
We split the data into 80\%~analysis and 20\%~test data. 
With the analysis data, we optimize classification pipelines using Naive AutoML~\cite{Mohr2022} (Sec.~\ref{sec:autoMLresults}) and evaluate the pipelines on the test data (Sec.~\ref{sec:autoMLtest}).
Next, we investigate the influence of the number of active robots (Sec.~\ref{sec:results_robots}) and the experimental duration (Sec.~\ref{sec:results_experimental_time}). 
Finally, we explore potential for fine tuning models to the individual user (Sec.~\ref{sec:results_individual}). 

\subsection{Optimizing ML pipelines for eye-tracking data}
\label{sec:autoMLresults}
To determine the optimal ML pipeline, we use the analysis dataset and maximize accuracy as the evaluation metric.  
%We optimize the machine learning pipeline using Naive AutoML~\cite{Mohr2022} and the metrics accuracy and ROC~AUC. 
%The two class problems are optimized using both metrics and the three class problem is only optimized for accuracy.
We optimize one ML pipeline for each time window size~$t_w$ for 2 or 3~classes, obtaining a total of 32~optimization settings. 
The optimization results using the \emph{duration estimate} label can be found in Table~\ref{tab:optimization_results_duration_estimate} and for the \emph{PPOT} label in Table~\ref{tab:optimization_results_ppot}. 
The exact hyperparameter choices can be found in the implementation. 
Throughout all different settings we see that the optimizer most often chose the \textit{RandomForestClassifier}~(RFC) or \textit{ExtraTreesClassifier}~(ETC) as classifiers~(CLF), while for few settings the \textit{KNeighborsClassifier}~(kNN) was chosen. 
The most often selected preprocessing step~(PRE) is \textit{VarianceThreshold}~(VT), followed by principal component analysis~(PCA), and \textit{Normalizer}~(NOR). 
No preprocessing step~(None) was chosen 11~times over the 32~optimizing settings.  
\begin{table}[]
\vspace{2pt}
\caption{Optimization results for the \emph{duration estimate} labeled analysis dataset.}
\label{tab:optimization_results_duration_estimate}
\centering
\begin{tabular}{@{}p{0.75cm}p{0.8cm}p{0.8cm}p{0.8cm}p{0.8cm}p{0.8cm}p{0.8cm}@{}}
\toprule
\multirow{2}{0.75cm}{$t_w$ [s]} & \multicolumn{3}{c}{2 classes} & \multicolumn{3}{c}{3 classes} \\ \cline{2-7}
                    & Pre & Clf & ACC & Pre & Clf & ACC\\ \midrule
1                   & None & ETC & 0.983 & None & ETC & 0.970 \\                    
2                   & None & RFC & \textbf{0.987}   & PCA  & RFC & \textbf{0.973}\\
5                   & VT   & RFC & 0.979   & None & RFC & 0.970\\
10                  & None & RFC & 0.953   & VT   & RFC & 0.949\\
15                  & None & RFC & 0.946   & VT   & RFC & 0.908\\
20                  & None & ETC   & 0.906 & VT   & ETC & 0.863  \\
30                  & VT   & ETC   & 0.877 & VT   & ETC & 0.836\\
45                  & VT   & ETC   & 0.843 & None & ETC & 0.789 \\
60                  & VT   & ETC   & 0.800 & VT   & ETC & 0.717\\ \bottomrule
\end{tabular}
\end{table}
\begin{table}[]
\caption{Optimization results for the \emph{PPOT} labeled analysis dataset.}
\label{tab:optimization_results_ppot}
\centering
\begin{tabular}{@{}p{0.75cm}p{0.8cm}p{0.8cm}p{0.8cm}p{0.8cm}p{0.8cm}p{0.8cm}@{}}
\toprule
\multirow{2}{0.75cm}{$t_w$ [s]} & \multicolumn{3}{c}{2 classes} & \multicolumn{3}{c}{3 classes} \\ \cline{2-7}
                    & Pre & Clf & ACC & Pre & Clf & ACC\\ \midrule
1                   & NOR & ETC & 0.973 & NOR & ETC & \textbf{0.975} \\ %\ta{+PF}
2                   & PCA  & kNN   & \textbf{0.974} & VT   & RFC & \textbf{0.975} \\
5                   & PCA  & kNN   & 0.953 & None & RFC & \textbf{0.975} \\
10                  & NOR  & ETC   & 0.923 & VT   & ETC   & 0.931\\
15                  & NOR  & ETC   & 0.908 & VT   & ETC   & 0.893\\
20                  & None & ETC   & 0.894 & VT   & ETC   & 0.853\\
30                  & None & ETC   & 0.870 & VT   & ETC   & 0.825\\
45                  & None & ETC   & 0.827 & VT   & ETC   & 0.769\\
60                  & None & ETC   & 0.791 & PCA  & ETC   & 0.710\\\bottomrule
\end{tabular}
\end{table}
%
%Converging to the same models indicates that these models allow for a good representation of the underlying dataset and suggests that the dataset is sufficiently large to reduce the impact of noise arising from random data splits and stochastic model initialization.
Given we converge to the same models, suggests that these models effectively capture the underlying dataset structure and that the dataset is sufficiently large to mitigate the impact of noise from random data splits and stochastic model initialization.
We observe that the better results are obtained using smaller time window sizes $t_w$, which could be due to the larger datasets. 
For a smaller time window size, we can extract more samples. 
The samples also have a higher correlation as the physiology of the human eye changes with finite speed.

\subsection{Evaluating optimized ML pipelines on the test dataset}
\label{sec:autoMLtest}
% see main.py
Next, we evaluate the optimized ML pipelines from the previous section using the unseen test dataset. 
We do 100~independent iterations, training with 80\% of the analysis data (stratified split) and all features, and then classify all test data samples.
The classification results of the test data using the \emph{duration estimate} and \emph{PPOT} label can be found in Table~\ref{tab:test_results}. 
%
% \begin{table}[]
% \caption{Classification accuracies of the unseen test data using the \emph{duration estimation} label ($\pm$ standard deviation).}
% \label{tab:test_duration_estimate}
% \centering
% \begin{tabular}{@{}p{0.75cm}p{2.0cm}p{2.0cm}@{}}
% \toprule
% $t_w$ [s] & 2 classes & 3 classes \\ \midrule
% 2	& 0.9852 $\pm$ 0.0015 & 0.9790 $\pm$ 0.0021 \\
% 5	& 0.9728 $\pm$ 0.0025 & 0.9705 $\pm$ 0.0019 \\
% 10	& 0.9549 $\pm$ 0.0051 & 0.9512 $\pm$ 0.0044 \\
% 15	& 0.9443 $\pm$ 0.0046 & 0.9360 $\pm$ 0.0067 \\
% 20	& 0.9172 $\pm$ 0.0062 & 0.8675 $\pm$ 0.0119 \\
% 30	& 0.8513 $\pm$ 0.0125 & 0.7782 $\pm$ 0.0152 \\
% 45	& 0.8644 $\pm$ 0.0181 & 0.7596 $\pm$ 0.0166 \\
% 60	& 0.8091 $\pm$ 0.0201 & 0.7099 $\pm$ 0.0263 \\ \bottomrule
% \end{tabular}
% \end{table}
%
% \begin{table}[]
% \caption{Classification accuracies of the unseen test data using the \emph{PPOT} label ($\pm$ standard deviation).}
% \label{tab:test_ppot}
% \centering
% \begin{tabular}{@{}p{0.75cm}p{2.4cm}p{2.4cm}@{}}
% \toprule
% $t_w$ [s] & 2 classes & 3 classes \\ \midrule
% 2	& 0.9720 $\pm$ 0.0020 & 0.9740 $\pm$ 0.0024 \\
% 5	& 0.9533 $\pm$ 0.0022 & 0.9641 $\pm$ 0.0027 \\
% 10	& 0.9191 $\pm$ 0.0040 & 0.9472 $\pm$ 0.0037 \\
% 15	& 0.9147 $\pm$ 0.0072 & 0.9178 $\pm$ 0.0064 \\
% 20	& 0.8800 $\pm$ 0.0075 & 0.8450 $\pm$ 0.0072 \\
% 30	& 0.8376 $\pm$ 0.0103 & 0.7804 $\pm$ 0.0141 \\
% 45	& 0.8335 $\pm$ 0.0151 & 0.7362 $\pm$ 0.0139 \\
% 60	& 0.8466 $\pm$ 0.0134 & 0.7010 $\pm$ 0.0222 \\\bottomrule
% \end{tabular}
% \end{table}
%
\begin{table}[]
\vspace{2pt}
\caption{Classification accuracies of unseen test data for \emph{duration estimation} and \emph{PPOT} labels ($\pm$~standard deviation).}
\label{tab:test_results}
\centering
\vspace{5pt}
\begin{tabular}{@{}p{0.5cm}>{\raggedright\arraybackslash}p{1.5cm}p{1.5cm}>{\raggedright\arraybackslash}p{1.5cm}p{1.5cm}@{}}
\toprule
\multirow{2}{0.75cm}{$t_w$ [s]} & \multicolumn{2}{c}{2 classes} & \multicolumn{2}{c}{3 classes} \\ \cline{2-5}
                    & duration estimate & PPOT & duration estimate & PPOT \\ \midrule
1   & 0.983$\pm$0.003 & 0.950$\pm$0.002 & 0.973$\pm$0.003 & 0.961$\pm$0.003 \\  
2	& \textbf{0.985$\pm$0.001} & \textbf{0.972$\pm$0.002} & \textbf{0.979$\pm$0.002} & \textbf{0.974$\pm$0.002} \\
5	& 0.972$\pm$0.002 & 0.953$\pm$0.002 & 0.970$\pm$0.001 & 0.964$\pm$0.002 \\
10	& 0.954$\pm$0.005 & 0.919$\pm$0.004 & 0.951$\pm$0.004 & 0.947$\pm$0.003 \\
15	& 0.944$\pm$0.004 & 0.914$\pm$0.007 & 0.936$\pm$0.006 & 0.917$\pm$0.006 \\
20	& 0.917$\pm$0.006 & 0.880$\pm$0.007 & 0.867$\pm$0.011 & 0.845$\pm$0.007 \\
30	& 0.851$\pm$0.012 & 0.837$\pm$0.010 & 0.778$\pm$0.015 & 0.780$\pm$0.014 \\
45	& 0.864$\pm$0.018 & 0.833$\pm$0.015 & 0.759$\pm$0.016 & 0.736$\pm$0.013 \\
60	& 0.809$\pm$0.020 & 0.846$\pm$0.013 & 0.709$\pm$0.026 & 0.701$\pm$0.022 \\\bottomrule
\end{tabular}
\vspace{-5pt}
\end{table}
The results on the unseen test data show that we are able to classify the subjective time perception, either measured through \emph{duration estimation} or \emph{PPOT}, based on the eye-tracking data with a high accuracy for all settings.
% make some statement about the performance 
The comparable performance, achieved using the full dataset and the analysis dataset, suggests that the models have effectively captured a robust representation of the underlying data structure.
% Discuss the influence of the time window 
Similar to the results using the analysis dataset to optimize the ML pipelines (Table~\ref{tab:optimization_results_duration_estimate} and Table~\ref{tab:optimization_results_ppot}), we observe that the classification performance increases for smaller time windows~$t_w$.
This can again be attributed to the larger datasets, which provide more training data that closely resemble the test data, as the temporal differences and consequently the variances are smaller.
This is because the human eye physiology changes in the range of milliseconds to seconds. 
Further, the performance decrease for larger $t_w$ is not explainable by variations in the class distribution, as for the two class datasets the variation is within 3\% (majority class using \emph{duration estimate}: 59\% for $t_w$~=~2~s, 56\% for $t_w$~=~60~s; majority class using \emph{PPOT}: 69\% for $t_w$~=~2~s, 66\% for $t_w$~=~60~s).
For the three class datasets, the variation is within 4\% (majority class using \emph{duration estimate}: 45\% for $t_w$~=~2~s, 41\% for $t_w$~=~60~s; majority class using \emph{PPOT}: 43\% for $t_w$~=~2~s, 44\% for $t_w$~=~60~s).
This is in contrast to~\cite{Guo2021} who found the 20~s time window to work best.

For our desired task, obtaining online feedback for a \textit{ChronoPilot} device, shorter time windows~$t_w$ are more desirable as feedback could be obtained faster and at higher frequency. 
Especially, if compared to other physiological signals used for human feedback, such as electrodermal activity or ECG, which require larger time windows or EEG, fMRI and fNIRs which are difficult to use under real world conditions, for example, working in a factory~\cite{Das2024}.

\subsection{Influence of the number of actively moving robots}
\label{sec:results_robots}
\begin{figure}[t]
    \centering
    \vspace{2pt}
    % \subfloat[Setting: 2 class \emph{duration estimate} label, optimized for accuracy.]{
    %     \includegraphics[width=0.45\textwidth]{figures/active_robots/2_classes/active_robots_accuracy_label_duration_estimate_bls.png}
    %     \label{fig:sub1}
    % }
    % \hspace{5pt}
    % \subfloat[Setting: 2 class \emph{PPOT} label, optimized for accuracy.]{
    %     \includegraphics[width=0.45\textwidth]{figures/active_robots/2_classes/active_robots_accuracy_label_ppot_bls.png}
    %     \label{fig:sub2}
    % }
    % \hfill
    % \subfloat[Setting: 3 class \emph{duration estimate} label, optimized for accuracy.]{
    %     \includegraphics[width=0.45\textwidth]{figures/active_robots/3_classes/active_robots_accuracy_label_duration_estimate_bls.png}
    %     \label{fig:sub5}
    % }
    % \hspace{5pt}
    % \subfloat[Setting: 3 class \emph{PPOT} label, optimized for accuracy.]{
    %     \includegraphics[width=0.45\textwidth]{figures/active_robots/3_classes/active_robots_accuracy_label_ppot_bls.png}
    %     \label{fig:sub6}
    % }
    %\includegraphics[width=0.45\textwidth]{figures/active_robots/2_classes/active_robots_accuracy_label_duration_estimate_bls.png}
    % This file was created with tikzplotlib v0.10.1.
\begin{tikzpicture}

\begin{axis}[
legend cell align={left},
legend columns=3,
legend style={
  fill opacity=0.8,
  draw opacity=1,
  text opacity=1,
  at={(0.11,0.03)},
  anchor=south west,
  draw=light-gray, 
  legend style={font=\scriptsize} % Change text size here
},
tick align=outside,
tick pos=left,
x grid style={dark-gray},
xlabel={Active robots [\#]},
xmin=-0.35, xmax=7.35,
xtick style={color=black},
xtick={0,1,2,3,4,5,6,7},
xticklabels={1,3,5,7,9,11,13,15},
y grid style={dark-gray},
ylabel={Accuracy},
ymin=0, ymax=1,
ytick style={color=black},
width=0.49\textwidth,
height=0.25\textwidth
]
\path [draw=blue, fill=blue, opacity=0.2]
(axis cs:0,1)
--(axis cs:0,0.988283289296296)
--(axis cs:1,0.990179737342349)
--(axis cs:2,0.988663389123865)
--(axis cs:3,0.985740826230299)
--(axis cs:4,0.970791018071116)
--(axis cs:5,0.98155182241638)
--(axis cs:6,0.987808269170184)
--(axis cs:7,0.981341996679763)
--(axis cs:7,1)
--(axis cs:7,1)
--(axis cs:6,1)
--(axis cs:5,1)
--(axis cs:4,1)
--(axis cs:3,1)
--(axis cs:2,1)
--(axis cs:1,1)
--(axis cs:0,1)
--cycle;

% \path [draw=blue, fill=blue, opacity=0.2]
% (axis cs:0,0.998988817070778)
% --(axis cs:0,0.956222450534856)
% --(axis cs:1,0.974603973854552)
% --(axis cs:2,0.971735544718073)
% --(axis cs:3,0.976128247046222)
% --(axis cs:4,0.972338050601291)
% --(axis cs:5,0.961808341359585)
% --(axis cs:6,0.967395356295406)
% --(axis cs:7,0.980684265241566)
% --(axis cs:7,1)
% --(axis cs:7,1)
% --(axis cs:6,1)
% --(axis cs:5,0.997080547529304)
% --(axis cs:4,0.999492935314201)
% --(axis cs:3,1)
% --(axis cs:2,0.997495224512697)
% --(axis cs:1,0.99961824836767)
% --(axis cs:0,0.998988817070778)
% --cycle;

\path [draw=blue, fill=blue, opacity=0.2]
(axis cs:0,1)
--(axis cs:0,0.953759556752643)
--(axis cs:1,0.967330578732818)
--(axis cs:2,0.962500948624989)
--(axis cs:3,0.96448553825298)
--(axis cs:4,0.952266974722178)
--(axis cs:5,0.937817470743773)
--(axis cs:6,0.9419897966466)
--(axis cs:7,0.964268032888358)
--(axis cs:7,1)
--(axis cs:7,1)
--(axis cs:6,0.996510203353399)
--(axis cs:5,0.992738084811782)
--(axis cs:4,0.999037373103909)
--(axis cs:3,1)
--(axis cs:2,0.998188706547425)
--(axis cs:1,1)
--(axis cs:0,1)
--cycle;

% \path [draw=blue, fill=blue, opacity=0.2]
% (axis cs:0,0.999386952896179)
% --(axis cs:0,0.945318929456762)
% --(axis cs:1,0.954202846012209)
% --(axis cs:2,0.963045203974039)
% --(axis cs:3,0.958839235688179)
% --(axis cs:4,0.929782200821046)
% --(axis cs:5,0.949018522362449)
% --(axis cs:6,0.90956853013107)
% --(axis cs:7,0.951379579355199)
% --(axis cs:7,1)
% --(axis cs:7,1)
% --(axis cs:6,0.995948711248241)
% --(axis cs:5,1)
% --(axis cs:4,0.994460223421378)
% --(axis cs:3,1)
% --(axis cs:2,0.998859557930723)
% --(axis cs:1,1)
% --(axis cs:0,0.999386952896179)
% --cycle;

\path [draw=blue, fill=blue, opacity=0.2]
(axis cs:0,0.993638112264896)
--(axis cs:0,0.913028554401771)
--(axis cs:1,0.940057092506841)
--(axis cs:2,0.953392111788756)
--(axis cs:3,0.958043069443811)
--(axis cs:4,0.926637418932449)
--(axis cs:5,0.940839194435388)
--(axis cs:6,0.924365762893718)
--(axis cs:7,0.916746405036563)
--(axis cs:7,1)
--(axis cs:7,1)
--(axis cs:6,1)
--(axis cs:5,1)
--(axis cs:4,1)
--(axis cs:3,1)
--(axis cs:2,1)
--(axis cs:1,1)
--(axis cs:0,0.993638112264896)
--cycle;

% \path [draw=blue, fill=blue, opacity=0.2]
% (axis cs:0,0.978823198868802)
% --(axis cs:0,0.852755748499619)
% --(axis cs:1,0.892126026844774)
% --(axis cs:2,0.912410673559336)
% --(axis cs:3,0.935755113179738)
% --(axis cs:4,0.873449643079439)
% --(axis cs:5,0.880560108885742)
% --(axis cs:6,0.870150162352193)
% --(axis cs:7,0.860281872476473)
% --(axis cs:7,0.987953421641174)
% --(axis cs:7,0.987953421641174)
% --(axis cs:6,0.996099837647807)
% --(axis cs:5,1)
% --(axis cs:4,0.996024041131088)
% --(axis cs:3,1)
% --(axis cs:2,1)
% --(axis cs:1,0.995873973155226)
% --(axis cs:0,0.978823198868802)
% --cycle;

\path [draw=blue, fill=blue, opacity=0.2]
(axis cs:0,0.962776515192028)
--(axis cs:0,0.797223484807972)
--(axis cs:1,0.825308805911662)
--(axis cs:2,0.809020658504515)
--(axis cs:3,0.882716851101372)
--(axis cs:4,0.782335951600328)
--(axis cs:5,0.764690547158785)
--(axis cs:6,0.724686937899625)
--(axis cs:7,0.691742886652474)
--(axis cs:7,0.893711658802072)
--(axis cs:7,0.893711658802072)
--(axis cs:6,0.917131243918557)
--(axis cs:5,0.953491271023033)
--(axis cs:4,0.951510202245826)
--(axis cs:3,1)
--(axis cs:2,0.967229341495485)
--(axis cs:1,0.973262622659767)
--(axis cs:0,0.962776515192028)
--cycle;

\path [draw=blue, fill=blue, opacity=0.2]
(axis cs:0,0.966363204695662)
--(axis cs:0,0.753636795304338)
--(axis cs:1,0.775705815282391)
--(axis cs:2,0.770417208843967)
--(axis cs:3,0.828783675849377)
--(axis cs:4,0.763520676947339)
--(axis cs:5,0.675793214046067)
--(axis cs:6,0.674071611530458)
--(axis cs:7,0.613671457718847)
--(axis cs:7,0.859661875614486)
--(axis cs:7,0.859661875614486)
--(axis cs:6,0.913428388469542)
--(axis cs:5,0.921706785953933)
--(axis cs:4,0.950479323052661)
--(axis cs:3,1)
--(axis cs:2,0.961249457822699)
--(axis cs:1,0.964294184717609)
--(axis cs:0,0.966363204695662)
--cycle;

\path [draw=red, fill=red, opacity=0.2]
(axis cs:0,0.597427221186668)
--(axis cs:0,0.526611168270361)
--(axis cs:1,0.545970814144264)
--(axis cs:2,0.498077378423274)
--(axis cs:3,0.502150597661737)
--(axis cs:4,0.6392853455159)
--(axis cs:5,0.58167224813635)
--(axis cs:6,0.63119382361983)
--(axis cs:7,0.567397546938562)
--(axis cs:7,0.633029637620501)
--(axis cs:7,0.633029637620501)
--(axis cs:6,0.659098746545303)
--(axis cs:5,0.648101978959306)
--(axis cs:4,0.688531814145701)
--(axis cs:3,0.57795574743598)
--(axis cs:2,0.529899517244776)
--(axis cs:1,0.569079413405963)
--(axis cs:0,0.597427221186668)
--cycle;

\addplot [semithick, blue, mark=*, mark size=3, mark options={solid}]
table {%
0 0.995862068965517
1 0.995811965811966
2 0.995454545454545
3 0.994262295081967
4 0.986
5 0.993469387755102
6 0.995652173913044
7 0.991805555555556
};
\addlegendentry{ACC (2 s)}
% \addplot [semithick, blue, dash pattern=on 5.55pt off 2.4pt, mark=*, mark size=3, mark options={solid}]
% table {%
% 0 0.977605633802817
% 1 0.987111111111111
% 2 0.984615384615385
% 3 0.989298245614035
% 4 0.985915492957746
% 5 0.979444444444444
% 6 0.984126984126984
% 7 0.991428571428572
% };
% \addlegendentry{ACC (5 s)}
\addplot [semithick, blue, dash pattern=on 1.5pt off 2.475pt, mark=*, mark size=3, mark options={solid}]
table {%
0 0.977234042553192
1 0.984545454545455
2 0.980344827586207
3 0.985555555555556
4 0.975652173913044
5 0.965277777777778
6 0.96925
7 0.9825
};
\addlegendentry{ACC (10 s)}
% \addplot [semithick, blue, dash pattern=on 9.6pt off 2.4pt on 1.5pt off 2.4pt, mark=*, mark size=3, mark options={solid}]
% table {%
% 0 0.97235294117647
% 1 0.978205128205128
% 2 0.980952380952381
% 3 0.981481481481481
% 4 0.962121212121212
% 5 0.978076923076923
% 6 0.952758620689655
% 7 0.979032258064516
% };
% \addlegendentry{ACC (15 s)}
\addplot [semithick, blue, dash pattern=on 7.5pt off 15pt, mark=*, mark size=3, mark options={solid}]
table {%
0 0.953333333333333
1 0.971
2 0.979375
3 0.986190476190476
4 0.963461538461539
5 0.974285714285714
6 0.966086956521739
7 0.960416666666667
};
\addlegendentry{ACC (20 s)}
% \addplot [semithick, blue, dash pattern=on 7.5pt off 1.5pt, mark=*, mark size=3, mark options={solid}]
% table {%
% 0 0.91578947368421
% 1 0.944
% 2 0.957391304347826
% 3 0.97625
% 4 0.934736842105263
% 5 0.946
% 6 0.933125
% 7 0.924117647058823
% };
% \addlegendentry{ACC (30 s)}
\addplot [semithick, blue, dash pattern=on 4.5pt off 15pt on 1.5pt off 15pt, mark=*, mark size=3, mark options={solid}]
table {%
0 0.88
1 0.899285714285714
2 0.888125
3 0.947
4 0.866923076923077
5 0.859090909090909
6 0.820909090909091
7 0.792727272727273
};
\addlegendentry{ACC (45 s)}
\addplot [semithick, blue, dash pattern=on 4.5pt off 1.5pt on 1.5pt off 1.5pt, mark=*, mark size=3, mark options={solid}]
table {%
0 0.86
1 0.87
2 0.865833333333333
3 0.92625
4 0.857
5 0.79875
6 0.79375
7 0.736666666666667
};
\addlegendentry{ACC (60 s)}
\addplot [semithick, red, dash pattern=on 5.55pt off 2.4pt, mark=*, mark size=3, mark options={solid}]
table {%
0 0.562019194728514
1 0.557525113775114
2 0.513988447834025
3 0.540053172548858
4 0.663908579830801
5 0.614887113547828
6 0.645146285082567
7 0.600213592279532
};
\addlegendentry{Majority class}
\end{axis}

\end{tikzpicture}
    \setlength{\abovecaptionskip}{-27pt} % Space above caption
    \setlength{\belowcaptionskip}{-10pt} % Space below caption
    \caption{Classification accuracies using the binary \emph{duration estimate} label for experiments split by actively moving robots. The blue lines indicate the accuracy of the classifier for different $t_w$ while the red line indicates the majority class of the subdatasets. The shaded areas represent the standard deviation (over 100~independent repetitions).\label{fig:individual_robots}}
\end{figure}
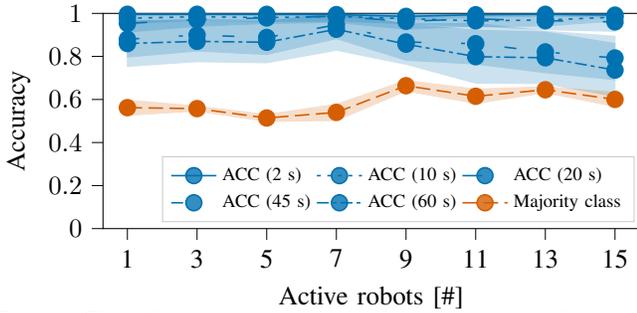

% see eye_tracking_robot.py
During our experimental procedure, one parameter we vary is the number of actively moving robots. 
The remaining (passive) robots stand still inside the arena to mitigate any side effects of varying the swarm size.
To mitigate that our approach classifies for the number of actively moving robots (see~\cite{Kaduk2024} for label distribution for varying active robots) we split the dataset by actively moving robots (e.g., using all experiments of $n$~actively moving robots). 
We use a stratified shuffle split and 100~independent repetitions. 
For space reasons we only present the results for binary \emph{duration estimate} labels in Fig.~\ref{fig:individual_robots}.
However, the other settings have qualitatively similar results. 
Fig.~\ref{fig:individual_robots} shows that the classification performance is consistent regardless of the number of active robots.  
Hence, we assume that the ML models do not classify for the number of robots. 

\subsection{Influence of the experiment duration}
\label{sec:results_experimental_time}
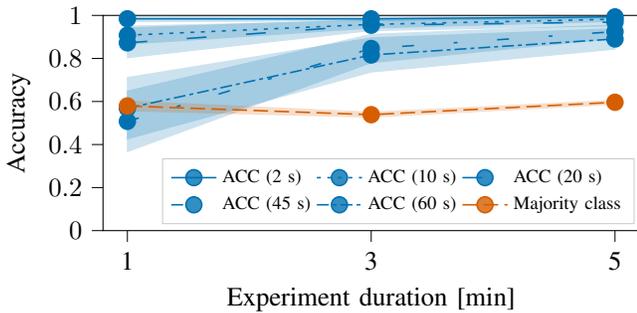
\begin{figure}[t]
    \centering
    % \subfloat[Setting: 2 class \emph{duration estimate} label, optimized for accuracy.]{
    %     \includegraphics[width=0.45\textwidth]{figures/experimental_time/2_classes/experimental_time_accuracy_label_duration_estimate_bls.png}
    %     \label{fig:sub1_1}
    % }
    % \hspace{5pt}
    % \subfloat[Setting: 2 class \emph{PPOT} label, optimized for accuracy.]{
    %     \includegraphics[width=0.45\textwidth]{figures/experimental_time/2_classes/experimental_time_accuracy_label_ppot_bls.png}
    %     \label{fig:sub2_1}
    % }
    % \hfill
    % \subfloat[Setting: 3 class \emph{duration estimate} label, optimized for accuracy.]{
    %     \includegraphics[width=0.45\textwidth]{figures/experimental_time/3_classes/experimental_time_accuracy_label_duration_estimate_bls.png}
    %     \label{fig:sub5_1}
    % }
    % \hspace{5pt}
    % \subfloat[Setting: 3 class \emph{PPOT} label, optimized for accuracy.]{
    %     \includegraphics[width=0.45\textwidth]{figures/experimental_time/3_classes/experimental_time_accuracy_label_ppot_bls.png}
    %     \label{fig:sub6_1}
    % }
    % \includegraphics[width=0.45\textwidth]{figures/experimental_time/2_classes/experimental_time_accuracy_label_duration_estimate_bls.png}
    % This file was created with tikzplotlib v0.10.1.
\begin{tikzpicture}

\begin{axis}[
legend cell align={left},
legend columns=3,
legend style={
  fill opacity=0.8,
  draw opacity=1,
  text opacity=1,
  at={(0.11,0.03)},
  anchor=south west,
  draw=light-gray,
  legend style={font=\scriptsize}
},
tick align=outside,
tick pos=left,
x grid style={dark-gray},
xlabel={Experiment duration [min]},
xmin=-0.1, xmax=2.1,
xtick style={color=black},
xtick={0,1,2},
xticklabels={1,3,5},
y grid style={dark-gray},
ylabel={Accuracy},
ymin=0, ymax=1,
ytick style={color=black},
width=0.49\textwidth,
height=0.25\textwidth
]
\path [draw=blue, fill=blue, opacity=0.2]
(axis cs:0,0.998257922125451)
--(axis cs:0,0.970313506445977)
--(axis cs:1,0.975595615744321)
--(axis cs:2,0.988285701889181)
--(axis cs:2,0.997738394496361)
--(axis cs:2,0.997738394496361)
--(axis cs:1,0.992440098541394)
--(axis cs:0,0.998257922125451)
--cycle;

% \path [draw=blue, fill=blue, opacity=0.2]
% (axis cs:0,0.982302908769022)
% --(axis cs:0,0.937995598693664)
% --(axis cs:1,0.967952345796579)
% --(axis cs:2,0.979821681727267)
% --(axis cs:2,0.992098848074058)
% --(axis cs:2,0.992098848074058)
% --(axis cs:1,0.989103491766872)
% --(axis cs:0,0.982302908769022)
% --cycle;

\path [draw=blue, fill=blue, opacity=0.2]
(axis cs:0,0.949916008208648)
--(axis cs:0,0.866181552766962)
--(axis cs:1,0.940936342419691)
--(axis cs:2,0.972356923929045)
--(axis cs:2,0.99225846068634)
--(axis cs:2,0.99225846068634)
--(axis cs:1,0.975543657580308)
--(axis cs:0,0.949916008208648)
--cycle;

% \path [draw=blue, fill=blue, opacity=0.2]
% (axis cs:0,0.962121120126508)
% --(axis cs:0,0.851212213206825)
% --(axis cs:1,0.936263325524884)
% --(axis cs:2,0.96614311668709)
% --(axis cs:2,0.991266955255356)
% --(axis cs:2,0.991266955255356)
% --(axis cs:1,0.976127978822942)
% --(axis cs:0,0.962121120126508)
% --cycle;

\path [draw=blue, fill=blue, opacity=0.2]
(axis cs:0,0.941770625043525)
--(axis cs:0,0.803446766260822)
--(axis cs:1,0.929342728204552)
--(axis cs:2,0.952016266008704)
--(axis cs:2,0.985391141398703)
--(axis cs:2,0.985391141398703)
--(axis cs:1,0.979800128938304)
--(axis cs:0,0.941770625043525)
--cycle;

% \path [draw=blue, fill=blue, opacity=0.2]
% (axis cs:0,0.857337550814528)
% --(axis cs:0,0.656412449185472)
% --(axis cs:1,0.882857442096702)
% --(axis cs:2,0.918423210695778)
% --(axis cs:2,0.967366262988433)
% --(axis cs:2,0.967366262988433)
% --(axis cs:1,0.948742557903298)
% --(axis cs:0,0.857337550814528)
% --cycle;

\path [draw=blue, fill=blue, opacity=0.2]
(axis cs:0,0.648372243674832)
--(axis cs:0,0.367183311880724)
--(axis cs:1,0.780229983045824)
--(axis cs:2,0.893792941552604)
--(axis cs:2,0.955466317706655)
--(axis cs:2,0.955466317706655)
--(axis cs:1,0.912122958130647)
--(axis cs:0,0.648372243674832)
--cycle;

\path [draw=blue, fill=blue, opacity=0.2]
(axis cs:0,0.71129961036124)
--(axis cs:0,0.42620038963876)
--(axis cs:1,0.736824730434579)
--(axis cs:2,0.843155212761016)
--(axis cs:2,0.937870428264625)
--(axis cs:2,0.937870428264625)
--(axis cs:1,0.895482961873114)
--(axis cs:0,0.71129961036124)
--cycle;

\path [draw=red, fill=red, opacity=0.2]
(axis cs:0,0.60174617677126)
--(axis cs:0,0.556796301028441)
--(axis cs:1,0.527390984874104)
--(axis cs:2,0.587227109292846)
--(axis cs:2,0.605521618545984)
--(axis cs:2,0.605521618545984)
--(axis cs:1,0.550131209899685)
--(axis cs:0,0.60174617677126)
--cycle;

\addplot [semithick, blue, mark=*, mark size=3, mark options={solid}]
table {%
0 0.984285714285714
1 0.984017857142857
2 0.993012048192771
};
\addlegendentry{ACC (2 s)}
% \addplot [semithick, blue, dash pattern=on 5.55pt off 2.4pt, mark=*, mark size=3, mark options={solid}]
% table {%
% 0 0.960149253731343
% 1 0.978527918781726
% 2 0.985960264900662
% };
% \addlegendentry{ACC (5 s)}
\addplot [semithick, blue, dash pattern=on 1.5pt off 2.475pt, mark=*, mark size=3, mark options={solid}]
table {%
0 0.908048780487805
1 0.95824
2 0.982307692307693
};
\addlegendentry{ACC (10 s)}
% \addplot [semithick, blue, dash pattern=on 9.6pt off 2.4pt on 1.5pt off 2.4pt, mark=*, mark size=3, mark options={solid}]
% table {%
% 0 0.906666666666667
% 1 0.956195652173913
% 2 0.978705035971223
% };
% \addlegendentry{ACC (15 s)}
\addplot [semithick, blue, dash pattern=on 7.5pt off 15pt, mark=*, mark size=3, mark options={solid}]
table {%
0 0.872608695652174
1 0.954571428571428
2 0.968703703703704
};
\addlegendentry{ACC (20 s)}
% \addplot [semithick, blue, dash pattern=on 7.5pt off 1.5pt, mark=*, mark size=3, mark options={solid}]
% table {%
% 0 0.756875
% 1 0.9158
% 2 0.942894736842105
% };
% \addlegendentry{ACC (30 s)}
\addplot [semithick, blue, dash pattern=on 4.5pt off 15pt on 1.5pt off 15pt, mark=*, mark size=3, mark options={solid}]
table {%
0 0.507777777777778
1 0.846176470588235
2 0.924629629629629
};
\addlegendentry{ACC (45 s)}
\addplot [semithick, blue, dash pattern=on 4.5pt off 1.5pt on 1.5pt off 1.5pt, mark=*, mark size=3, mark options={solid}]
table {%
0 0.56875
1 0.816153846153846
2 0.890512820512821
};
\addlegendentry{ACC (60 s)}
\addplot [semithick, red, dash pattern=on 5.55pt off 2.4pt, mark=*, mark size=3, mark options={solid}]
table {%
0 0.57927123889985
1 0.538761097386895
2 0.596374363919415
};
\addlegendentry{Majority class}
\end{axis}

\end{tikzpicture}
    \setlength{\abovecaptionskip}{-27pt} % Space above caption
    \setlength{\belowcaptionskip}{-5pt} % Space below caption
    \caption{Classification accuracies using the binary \emph{duration estimate} label for experiments split by experiment duration. The blue lines indicate the accuracy of the classifier for different $t_w$ while the red line indicates the majority class of the subdatasets. The shaded areas represent the standard deviation (over 100~independent repetitions).\label{fig:experimental_time}}
\end{figure}

% see eye_tracking_time.py
The second parameter that is varied in the experiments is the experiment duration. 
To investigate if the experiment duration has an influence on the classification results we test our ML pipelines using only one experiment duration at a time. 
We use a stratified shuffle split and 100~independent repetitions. 
Again, for space reasons we only show the results obtained using the binary \emph{duration estimate} labels in Fig.~\ref{fig:experimental_time} (the other settings are qualitatively similar). 
Using only the 1~minute experiments we can see a dramatic decrease in performance for $t_w$~=~\{45~s, 60~s\}. 
This can be explained by the small dataset, as for both parameters only one sample per experiment can be generated. 
If we allow for more samples, for example, in 3~minute and 5~minute experiments, the performance goes back to normal. 
Similar, to the number of actively moving robots we can assume that the ML models do not classify for the experiment duration. 

\subsection{Fine tuning for the individual user}
\label{sec:results_individual}
% see eye_tracking_participants.py
\begin{table}[h]
    \vspace{5pt}
    \caption{The values are in percentage points and show by how much the fine-tuned classifier outperforms the not fine-tuned classifier.}
    \label{tab:participants_results}
    \centering
    \begin{tabular}{c|ccccccccc}
    \toprule
    & \multicolumn{9}{c}{$t_w$}\\
    P\_ID & 1 & 2 & 5 & 10 & 15 & 20 & 30 & 45 & 60 \\\midrule
% 2  & 21 & 19 & 20 & 22 & 25 & 23 & \textbf{24} & \textbf{28} \\
% 3  & 21 & 11 & 09 & 12 & 17 & 12 &  8 & 16 \\
% 4  & 25 & 20 & 14 & 11 & 23 & 17 & 20 & \textbf{28} \\
% 5  & 31 & 36 & 20 & 28 & 23 & \textbf{24} & 22 & 24 \\
% 6  & 47 & \textbf{52} & \textbf{39} & \textbf{36} & \textbf{29} & 13 & 22 & 21 \\
% 10 & 17 & 20 & 15 & 16 &  9 &  6 &  6 &  9 \\
% 11 & 13 & 10 & 12 &  7 & 13 &  9 & 18 & 19 \\
% 12 & 10 &  7 &  4 &  4 &  1 &  4 &  6 & 13 \\
% 13 & 13 & 18 & 13 &  6 &  6 &  7 & 10 &  9 \\
% 14 &  0 &  3 & 11 & 14 & 24 & 19 & 17 &  7 \\
% 15 & 37 & 22 & 17 & 16 & 16 & 12 & 11 & 17 \\
% 16 & 34 & 27 & 21 & 21 & 16 & 10 & 12 & 10 \\
% 17 & \textbf{69} & 32 & 33 & \textbf{36} & 25 & 18 & 13 & 11 \\
% 18 & 23 & 13 & 22 & 25 & 27 & 12 & 12 & 15 \\
% 21 & 51 & 43 & 35 & 34 & 23 & 13 & 10 & 12 \\
% 22 & 33 & 26 & 12 & 14 &  6 &  5 &  8 & 12 \\
% 23 & 30 & 19 & 15 &  9 &  7 &  7 & 10 &  6 \\\midrule
% MEAN &28 & 22 & 18 & 18 & 17 & 12 & 13 & 15 \\
2 & 	5 & 	20 & 	18 & 	20 & 	21 & 	25 & 	\textbf{23} & 	\textbf{23} & 	27 \\ 
3 & 	7 & 	21 & 	10 & 	8 & 	12 & 	17 & 	12 & 	7 & 	15 \\ 
4 & 	20 & 	24 & 	19 & 	13 & 	11 & 	22 & 	16 & 	20 & 	\textbf{28} \\ 
5 & 	6 & 	30 & 	36 & 	19 & 	28 & 	22 & 	\textbf{23} & 	21 & 	24 \\ 
6 &     40 & 	47 & 	\textbf{52} & 	\textbf{39} & 	\textbf{36} & 	\textbf{29} & 	12 & 	22 & 	20 \\
10 & 	21 & 	17 & 	19 & 	14 & 	15 & 	8 & 	6 & 	6 & 	8 \\
11 & 	10 & 	12 & 	9 & 	12 & 	7 & 	13 & 	8 & 	18 & 	18 \\ 
12 & 	4 & 	9 & 	7 & 	4 & 	4 & 	0 & 	3 & 	5 & 	13 \\ 
13 & 	10 & 	12 & 	17 & 	13 & 	6 & 	6 & 	6 & 	9 & 	8 \\ 
14 & 	-1 & 	0 & 	2 & 	10 & 	13 & 	24 & 	18 & 	17 & 	7 \\ 
15 & 	33 & 	36 & 	22 & 	17 & 	16 & 	16 & 	11 & 	11 & 	17 \\ 
16 & 	29 & 	33 & 	27 & 	20 & 	20 & 	15 & 	10 & 	11 & 	9 \\ 
17 & 	37 & 	\textbf{69} & 	31 & 	32 & 	35 & 	24 & 	17 & 	12 & 	11 \\ 
18 & 	3 & 	23 & 	12 & 	21 & 	25 & 	27 & 	12 & 	12 & 	14 \\ 
21 & 	\textbf{55} & 	50 & 	43 & 	34 & 	33 & 	22 & 	13 & 	9 & 	11 \\ 
22 & 	22 & 	32 & 	26 & 	11 & 	13 & 	6 & 	5 & 	7 & 	12 \\ 
23 & 	22 & 	30 & 	19 & 	14 & 	9 & 	7 & 	6 & 	9 & 	6 \\\midrule
MEAN & 19 & 27 & 22 & 18 & 18 & 17 & 12 & 13 & 15 \\
        \bottomrule
    \end{tabular}
\end{table}
For our future application of delivering human feedback to a \textit{ChronoPilot} device, we can assume a ``setup" phase that allows to collect some initial data of the current user to fine tune our models to that individual. 
Here, we investigate if we can fine tune the models for individual users. 
We use the data of the first 30~seconds of all experiments from one user combined with all data from all other users to train or adjust classifiers to this particular user. 
We use the remaining data of this user (e.g., 30~s from the 1~min experiment, 2:30~min of the 3~min experiment, 4:30~min of the the 5~min experiment) to evaluate the classifiers. 
We compare this approach to using all data of one user for testing while training on the remaining data. 
See Table~\ref{tab:participants_results} for the averaged results over 100~independent repetitions of the improvement of accuracy by fine tuning the model for individual users for the binary \emph{duration estimate} label. 
Depending on the user and setting, we can substantially increase the performance (up to~69\%). 
We conclude that an initial ``setup" phase can massively improve the user feedback for a \textit{ChronoPilot} device.

\section{Conclusion}
% We are able to show that eye-tracking data can be effectively used to predict human subjective time perception. 
We have shown that we can process eye-tracking data combined with an automated ML approach to provide input to a future \textit{ChronoPilot} device.
The device would then decide autonomously over actions to modulate subject time perception. 
By utilizing automated ML methods we are able to find fine-tuned ML pipelines that are able to capture the properties of the underlying data. 
We mitigated the risk of classifiers relying on experiment-specific parameters, such as the number of actively moving robots or experiment duration. 
In our application scenario of providing feedback for a \textit{ChronoPilot} device, we demonstrate that incorporating a brief  ``setup" phase to adjust the model to the individual user yields significant benefits. 
%In future work we will integrate these findings into a closed-loop control approach to actively steer the subjective time perception of a human operator that controls a swarm of robots. %\ta{TODO: @JK this should introduce the final experiment we have planned together}

By integrating our conceptional framework of the \textit{ChronoPilot} device into general adaptive robotic control strategies, robot systems could dynamically adjust task complexity and interaction modalities to enhance operator performance and well-being. 
Future research could explore the integration of additional physiological signals with eye-tracking, incorporate multimodal feedback mechanisms, and apply our method to a broader range of collaborative human-robot scenarios. 
We hope to contribute to the vision of intelligent large-scale multi-robot systems that are not only efficient but also human-aware, enabling safer, more scalable, and more user-centered automation in high-demand environments.

\addtolength{\textheight}{-12cm}   % This command serves to balance the column lengths

%%%%%%%%%%%%%%%%%%%%%%%%%%%%%%%%%%%%%%%%%%%%%%%%%%%%%%%%%%%%%%%%%%%%%%%%%%%%%%%%

\section*{Acknowledgment}

This work has been partially supported by the European Union's Horizon 2020 FET research program under grant agreement 964464 (\textit{ChronoPilot}) and the DFG under Germany's Excellence Strategy, EXC 2117, 422037984 (H.H.).

\bibliographystyle{IEEEtran}
\bibliography{IEEEabrv,root}

\end{document}